\documentclass[conference]{IEEEtran}
\IEEEoverridecommandlockouts
\usepackage{cite}
\usepackage{amsmath,amssymb,amsfonts}
\usepackage{algorithmic}
\usepackage{multirow}
\usepackage{threeparttable}
\usepackage{graphicx}
\usepackage{textcomp}
\usepackage{xcolor}
\usepackage[latin1]{inputenc}

\def\BibTeX{{\rm B\kern-.05em{\sc i\kern-.025em b}\kern-.08em
    T\kern-.1667em\lower.7ex\hbox{E}\kern-.125emX}}
\begin{document}

\title{Diagonal Hierarchical Consistency Learning for Semi-supervised Medical Image Segmentation
}

\author{\IEEEauthorblockN{Heejoon Koo$^{\dagger}$} \thanks{$^{\dagger}$Department of Electronic and Electrical Engineering, University College London, London, United Kindom. Email: heejoon.koo.17@alumni.ucl.ac.uk}}

\maketitle

\begin{abstract}
Medical image segmentation, which is essential for many clinical applications, has achieved almost human-level performance via data-driven deep learning technologies. Nevertheless, its performance is predicated upon the costly process of manually annotating a vast amount of medical images. To this end, we propose a novel framework for robust semi-supervised medical image segmentation using diagonal hierarchical consistency learning (DiHC-Net). First, it is composed of multiple sub-models with identical multi-scale architecture but with distinct sub-layers, such as up-sampling and normalisation layers. Second, with mutual consistency, a novel consistency regularisation is enforced between one model's intermediate and final prediction and soft pseudo labels from other models in a diagonal hierarchical fashion. A series of experiments verifies the efficacy of our simple framework, outperforming all previous approaches on public benchmark dataset covering organ and tumour. \\
\end{abstract}


\section{Introduction}
Medical image segmentation (MIS) involves assigning each pixel in an image to a specific class of anatomical structure, such as organs, tissues and tumours. This task provides valuable information for both computer-aided diagnosis (CAD) and computer-assisted surgery (CAS) in intuitive and efficient manner. Recently, deep learning with U-Net-based architectures \cite{ronneberger2015u, milletari2016v} has significantly improved MIS. Nonetheless, acquiring a large volume of high-quality annotated medical images remains labour-intensive and time-consuming, necessitating medical experts to manually annotate intricate medical images. To address this issue, many studies have endeavoured on semi-supervised learning that uses a small set of labelled data in conjunction with a large set of unlabelled data \cite{jiao2022learning}. 

Accordingly, semi-supervised medical image segmentation (SSMIS) has undergone significant advancements. After a seminal work by \cite{yu2019uncertainty}, \cite{li2021hierarchical} extends it by leveraging deep supervision and hierarchical consistency regularisation. Meanwhile, \cite{li2020shape} predicts Signed Distance Maps (SDM) \cite{ma2020distance} as an auxiliary task in an adversarial manner \cite{goodfellow2014generative} to produce more realistic predictions. Thereafter, \cite{zhang2021dual} introduces dual task consistency between the SDM regression and segmentation prediction. 

Another line of literature \cite{yu2019uncertainty, wu2022mutual, huang2023complementary} introduces uncertainty estimation. Yu et al. \cite{yu2019uncertainty} introduces uncertainty maps obtained via Monte-Carlo (MC) Dropout \cite{gal2016dropout} to predict with higher confidence in a teacher-student framework. On the other hand, based on finding \cite{lakshminarayanan2017simple} that uncertainty can be quantified by a statistical discrepancy from multiple model predictions, \cite{wu2022mutual, huang2023complementary} design a network with multiple distinct sub-models. 

During the advancement of SSMIS, AI research community has delved into normalisation layers. Instance normalisation \cite{ulyanov2016instance} is developed for style transfer, while group normalisation \cite{wu2018group} has shown great stability on relatively small batch sizes. Albeit these findings, the utilisation of various normalisation layers in SSMIS still remains under-investigated \cite{zhou2019normalization}.

Motivated by recent advancements, we introduce a novel SSMIS framework under the assumption that a network, composed of diversified sub-models, can first fully learn from the scarce labelled data then collaborate by minimising disparities in predictions on uncertain regions yielded from both labelled and unlabelled data. Therefore, the main contributions of this paper is threefold. First, we design the network to be comprised of three identical multi-scale V-Nets \cite{milletari2016v} with distinct sub-layers, such as up-sampling and normalisation layers to increase intra-model diversity. Second, it is optimised via deep supervision on labelled data and our proposed diagonal hierarchical consistency learning as well as mutual consistency learning \cite{wu2022mutual} on both labelled and unlabelled data. Finally, this simple yet effective framework demonstrates its superiority by surpassing existing baselines on publicly available dataset of Left Atrium (LA) \cite{xiong2021global} and Brain Tumor Segmentation (BraTS) 2019 dataset \cite{brats2015}. 

\section{Methodologies}
\label{sec:methodologies}

\subsection{Task Formulation}
\label{sec:taskformulation}

In SSMIS, a set of training data has $N$ labelled data and $M$ unlabelled data, where $ N \ll M $. Then, the small set of labelled data is denoted as $ \mathfrak{D}_L = \{ x_i^l, y_i \}_{i=1}^N $ and remaining set of unlabelled data as $ \mathfrak{D}_U = \{ x_i^u \}_{i=1}^M $, where $ x_i^l $ and $ x_i^u $ denotes the input images and $ y_i $ the corresponding segmentation masks. The objective is to train a model $f_{\theta}$, parameterised with $ \theta $, using the full data $ \mathfrak{D} = \mathfrak{D}_L \cup \mathfrak{D}_U $ to perform pixel-wise classification correctly.

\begin{figure*}[htbp]
    \centering
    \includegraphics[width=0.65\textwidth]{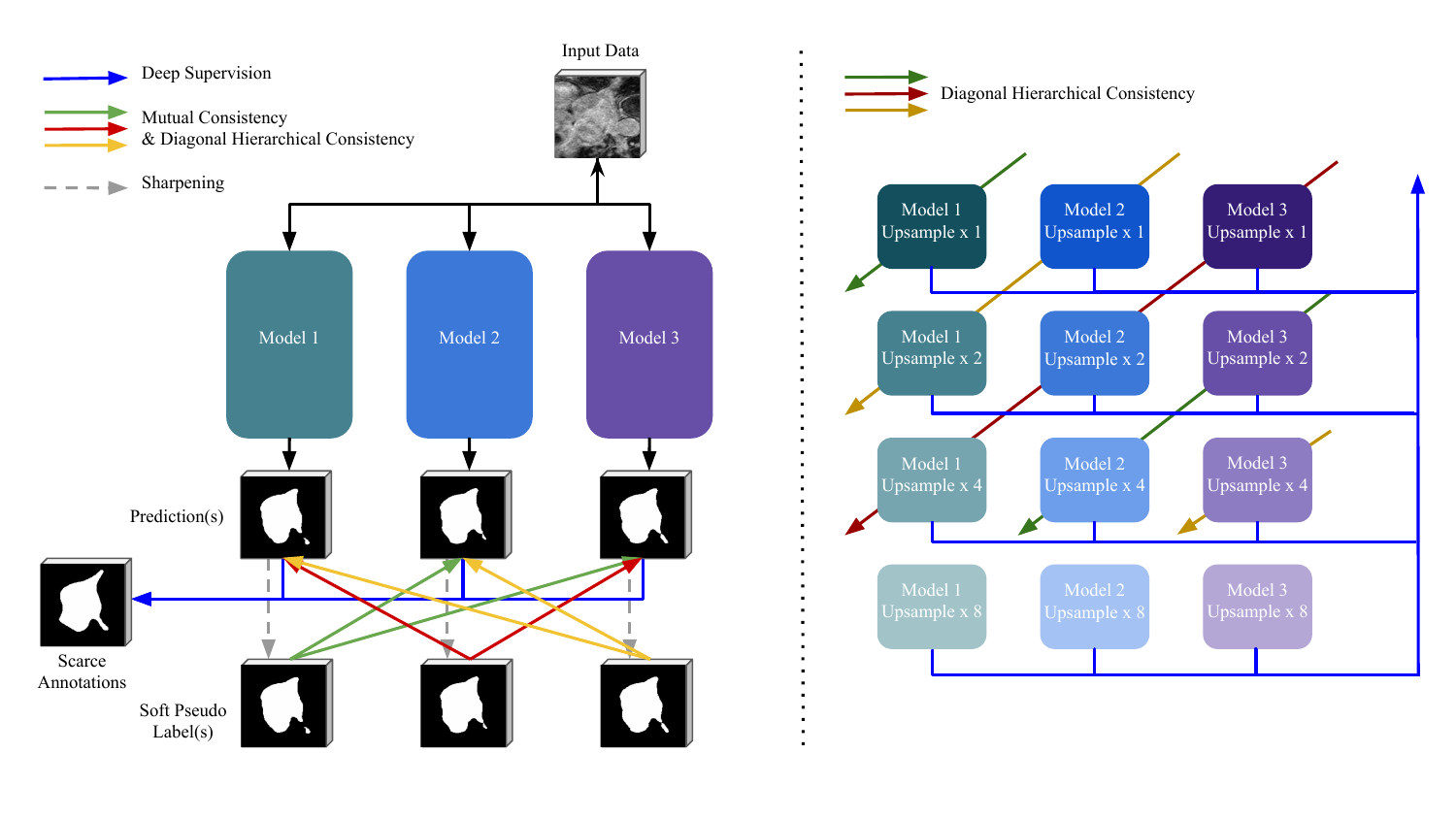}
    \caption{An Overview of Our Proposed Framework, DiHC-Net. The left provides a high-level conceptualisation of the proposed framework, whilst the right presents a visualisation on both deep-supervision and diagonal hierarchical consistency.}
    \label{figure_main}
\end{figure*}

\subsection{Diversifying Multi-scale Sub-models and Deep Supervision}
\label{sec:networkarchitecture}

Following the previous works \cite{wu2022mutual, huang2023complementary}, we employ a network with three sub-models having the identical architecture but with a variance in their sub-layers. Specifically, we use group normalisation \cite{wu2018group} and linear interpolation for the first sub-model's normalisation and up-sampling, respectively. The second sub-model configures with transposed convolution and batch normalisation \cite{ioffe2015batch} and the third model adopts nearest interpolation and instance normalisation \cite{ulyanov2016instance}.

Also, we extend our sub-models to have multi-scale architecture to produce intermediate representations from each block in their decoding stage. On the labelled data, each sub-model in the network learns scarce supervisory signal effectively by minimising the differences between up-sampled intermediate predictions and ground truths. This supervised segmentation loss is mathematically formulated as:
\begin{equation}
    \mathcal{L}_{sup} = \Sigma_{m=1}^{M} \Sigma_{s=1}^{S} \mathcal{L}_{dice} (f^{s}_{m}(x_{i}^{l}), y_{i}),
\end{equation}

where $ \mathcal{L}_{dice} $ represent dice loss, $ f^{s}_{m}(\cdot) $ denotes the predicted results from at scale $ s $ of $ m $-th model. Both $ s $ and and $ m $ are integers, ranging from 1 to 3 and 4, respectively.

\subsection{Diagonal Hierarchical Consistency Learning}
\label{sec:diagonalhierarchicalconsistency}

Boosting diversity amongst sub-models leads to increased inconsistent predictions especially in challenging regions. Pseudo labels have been effectively used to reduce uncertainties by minimizing entropy across models, with prior research employing mutual consistency loss to achieve this \cite{wu2022mutual, huang2023complementary}. However, they do not consider intermediate outputs. Addressing this gap, we introduce our novel diagonal hierarchical consistency loss. We use two consistency losses to both labelled and unlabelled data. 

First, we enforce mutual consistency between the soft pseudo labels converted from one sub-model's final prediction and others' final predictions. The sharpening function \cite{xie2020unsupervised} is utilised to acquire the soft pseudo label as follows:

\begin{equation}
\begin{aligned}
    S^{M_n} &= \text{sharpening}(p^{M_n}), \\
    \text{where} \hspace*{1mm} \text{sharpening} &= \frac{p^{1/T}}{p^{1/T} + (1-p)^{(1/T)}}.
\end{aligned}
\end{equation}

Here, $ S $ and $ p $ represent the pseudo label and prediction, respectively, while $ M_n $ denotes the $ n $-th model. The hyper-parameter $ T $ serves to control entropy; a lower $ T $ promotes low-entropy predictions. Then, with $ \alpha_{k} $ as coefficient for the $ k $-th block at the reverse decoding order, formulation of the mutual consistency loss using mean squared error (MSE) is:

\begin{equation}
    \mathcal{L}_{mc} = \Sigma_{i, j=1 \& i \neq j}^{n} \alpha_{1} \text{MSE}(S_{i}^{M_i}, p_{j}^{M_j}).
\end{equation}

\begin{table*}[!htb]
	\centering
	\caption{A Quantitative Performance Comparison on the LA and BraTS Dataset.}
        \begin{threeparttable}
	\resizebox{1.0\textwidth}{!}{
	\begin{tabular}{c|cc|cccc|cccc}
		\hline 
		\hline
		\multirow{2}{*}{Method}&\multicolumn{2}{c}{\% Scans Used}&\multicolumn{4}{|c}{LA}&\multicolumn{4}{|c}{BraTS}\\
		\cline{2-11}
		&Labelled&Unlabelled &Dice(\%)$\uparrow$ &Jaccard(\%)$\uparrow$&ASD(voxel)$\downarrow$&95HD(voxel)$\downarrow$&Dice(\%)$\uparrow$ &Jaccard(\%)$\uparrow$&ASD(voxel)$\downarrow$&95HD(voxel)$\downarrow$\\
		\hline
		\hline
		UA-MT \cite{yu2019uncertainty}& \multirow{6}{*}{10\%} &\multirow{6}{*}{90\%} &84.25 &73.48 &3.36&13.84 &80.85 &70.32&\textbf{2.57} &14.61  \\
		SASS \cite{li2020shape}  &  & &86.81 &76.92 &3.94 &12.54 & 81.96 & 70.65 & 4.92 & 16.99 \\
        DTC \cite{zhang2021dual}  &  & &87.51 &78.17 &2.36 &8.23 &81.96 &71.84  &2.43 &12.08 \\
		MC-Net+ \cite{wu2022mutual} & & &88.96 &80.25  &1.86 &7.93&82.42 & 72.44 & 4.38 & 13.94 \\
		CC-Net \cite{huang2023complementary} & & &89.82 &81.60 &1.8 &7.03  & 82.74 & 72.82 & 3.02 & 12.29    \\
		\textbf{Ours} (DiHC-Net) & & & \textbf{90.42} & \textbf{82.58} & \textbf{1.54} & \textbf{6.52}  & \textbf{84.96} & \textbf{75.36} & \textbf{2.57} & \textbf{10.33} \\
		\hline
		UA-MT \cite{yu2019uncertainty}& \multirow{6}{*}{20\%} &\multirow{6}{*}{80\%} &88.88 &80.21 &2.26 &7.32 &81.57 &71.42&2.49 &13.98 \\
		SASS \cite{li2020shape}  &  & &89.27 &80.82&3.13 &8.83 & 80.79 & 70.56 & 4.10& 13.80   \\
        DTC \cite{zhang2021dual}   &  & &89.42 &80.98 &2.1&7.32 & 82.78 & 72.47 & 2.2& 13.43  \\
		MC-Net+ \cite{wu2022mutual} & & &91.07 &83.67&1.67 &5.84  & 83.24 & 73.50& 2.55 & 9.78 \\
		CC-Net \cite{huang2023complementary} & & &91.27 &84.02 &1.54&5.75  & 83.30 & 73.34 & 2.57 & 10.44 \\
		\textbf{Ours} (DiHC-Net) & & & \textbf{91.94} & \textbf{85.13} & \textbf{1.35} & \textbf{5.03} & \textbf{85.47} & \textbf{76.13} & \textbf{2.16}& \textbf{9.31} \\
		\hline
	\end{tabular}}
	\label{table_quantitative_results}
    \end{threeparttable}
\end{table*}

Next, our novel diagonal hierarchical consistency minimises the difference between one sub-model's pseudo labels converted from its final representation and others' intermediate and last representations. We design it to be computationally efficient and not redundant, enabling the mutual regularisation of predictions across all sub-models. 

Let the coordinate of representation as ($m$-th model, $s$-th representation). Then, the representation from the last decoding block of the first model is (1, 1). Similarly, the representation from the first decoding block of the third model is (3, 4). Self-constraint is enforced between the pseudo label and the first decoder block from the same model if following our consistency strategy. However, we do not conduct it due to empirical harmful effects. Thus, the regarding loss is:
\begin{equation}
\begin{aligned}
    \mathcal{L}_{dihc}
    &= \alpha_{2} \text{MSE}(S_{1}^{M_1}, p_{2}^{M_3}) + \alpha_{3} \text{MSE}(S_{1}^{M_1}, p_{3}^{M_2}) \\
    &+ \alpha_{2} \text{MSE}(S_{2}^{M_2}, p_{2}^{M_1}) + \alpha_{3} \text{MSE}(S_{2}^{M_2}, p_{3}^{M_3}) \\
    &+ \alpha_{2} \text{MSE}(S_{3}^{M_3}, p_{2}^{M_2}) + \alpha_{3} \text{MSE}(S_{3}^{M_3}, p_{3}^{M_1}).
\end{aligned}
\end{equation}

Therefore, the network is optimised via a weighted sum of supervised loss $ \mathcal{L}_{sup} $ and consistency loss $ \mathcal{L}_{cst} $.
\begin{equation}
\begin{aligned}
  \mathcal{L}_{total} &= \lambda_{sup} \mathcal{L}_{sup} + \lambda_{cst} \mathcal{L}_{cst}, \\
  \text{where} \hspace*{1mm} \mathcal{L}_{cst} &= \mathcal{L}_{mc} + \mathcal{L}_{dihc},
\end{aligned}
\end{equation}

where $ \lambda_{sup} $ denotes a hyper-parameter for the supervised loss and $ \lambda_{cst} $ is the coefficient for the consistency loss, which is a time-dependent Gaussian warming-up function that balances between two loss terms. It is formalised as $ \lambda(t) = 0.1 \ast \text{exp}{(-5(1-t/t_{max}))} $ where $ t $ and $ t_{max} $ are present training step and the maximum step, respectively. It addresses the potential issue of consistency loss term that may harm the learning in the early stage.

\section{Experimental Setting}
\label{sec:experimentalsetting}

\subsection{Dataset}
To validate the effectiveness of our approach, we conduct experiments on publicly available LA and BraTS 2019 dataset. They provide total 100 and 335 3D MRI scans and corresponding segmentation masks. As test set annotations are publicly unavailable for LA dataset, the training set is divided into two subsets with 80 scans for training and 20 scans for evaluation. BraTS dataset is originally split into 250 scans for training, 25 scans for validation and the remaining 60 scans for testing. 

\subsection{Implementation Details} 
We follow the previous works \cite{yu2019uncertainty, li2020shape, wu2022mutual} by employing the same data pre-processing, data augmentation techniques, hyper-parameter and optimisation settings for training. Additionally, we do not use any post-processing module (e.g. non-maximum suppression) for a fair comparison. 

Soft pseudo labels are generated using a sharpening function with a fixed temperature $T$ of 0.1. We use three intermediate representations and coefficients for each consistency loss are determined as $\alpha_1$ = 1.0, $\alpha_2$ = 0.75, and $\alpha_3$ = 0.5. We set them to gradually decrease, since intermediate blocks should be softly regularised due to their tendency to produce relatively incomplete outputs. Also, we set $ \lambda_{sup} $ as 1. 

We experiment on two settings: training with 10\% or 20\% labelled data and the remaining unlabelled data. Instead of averaging three outputs, we utilise results from the first model, in line with prior research \cite{wu2022mutual, huang2023complementary}. We measure performance using four metrics: Dice, Jaccard, ASD (Average Surface Distance) and 95HD (95\% Hausdorff Distance). The first two metrics assess the interior segmentation predictions, whilst the last two assess the contour predictions. 

\begin{figure*}[ht!]
    \centering
    \includegraphics[width=0.8\textwidth]{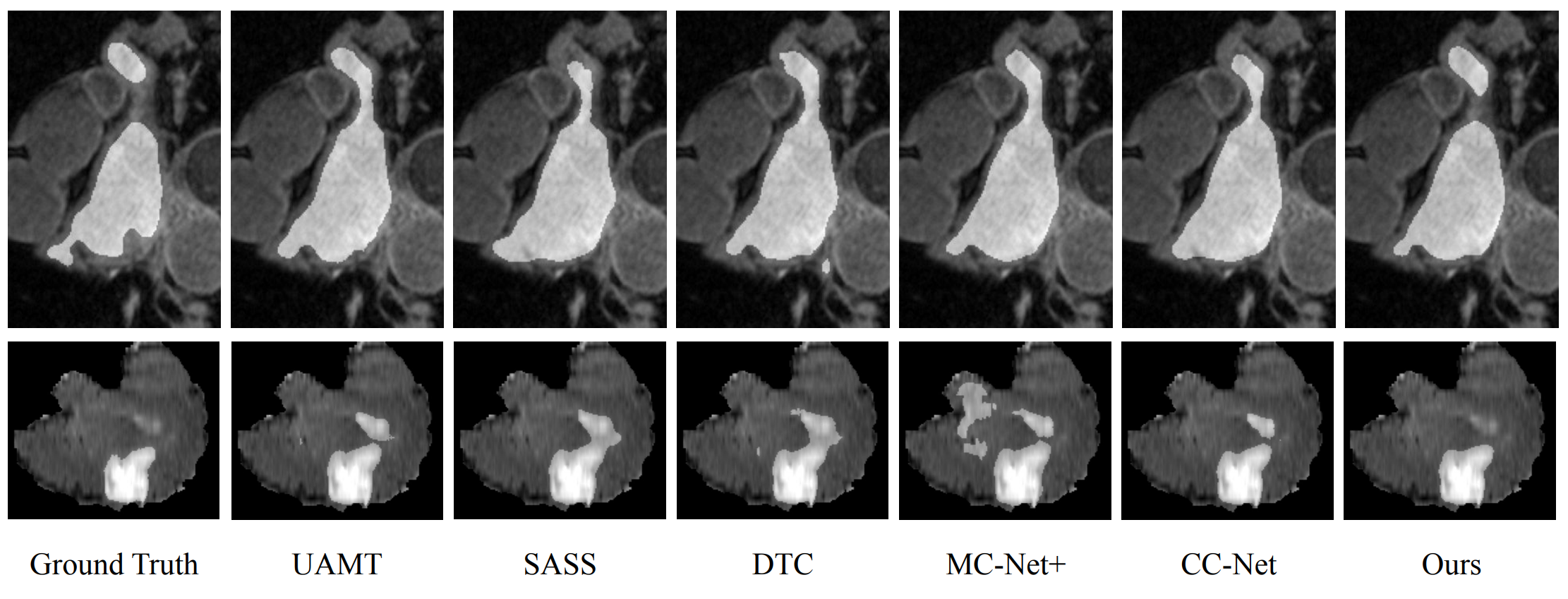}
    \caption{2D Visualisation of Ground Truth and Predictions of Our DiHC-Net and Other Baselines on LA and BraTS Dataset. The first row is acquired when trained with 10\% labelled data from LA dataset and the second row with 20\% labelled data from BraTS dataset.}
    \label{figure_qualitative_results}
\end{figure*}

\section{Experimental Results}
\label{sec:experimentalresults}

\subsection{Performance Comparison}
\label{sec:laresults}

Tab. \ref{table_quantitative_results} presents the quantitative results of our proposed method and other strong baselines, including UA-MT \cite{yu2019uncertainty}, SASS \cite{li2020shape}, DTC \cite{zhang2021dual}, MC-Net+ \cite{wu2022mutual} and CC-Net \cite{huang2023complementary}. As shown in Tab. \ref{table_quantitative_results}, our DiHC-Net consistently outperforms all baselines by a large margin consistently in both training settings across both dataset. Fig. \ref{figure_qualitative_results} visually highlights the effectiveness and robustness of our approach. Our framework produces more natural contours than shape-constraint methods \cite{li2020shape, zhang2021dual} and is more stable to false negatives on both dataset.

This superior performance is achieved by multiple sub-models that first fully learn scarce labelled data then reduce discrepancies in ambiguous and challenging areas from one model's intermediate and final predictions and other's pseudo labels by employing a our novel diagonal hierarchical consistency approach in all data.

\subsection{Ablation Studies}
\label{sec:ablationstudies}

\begin{table}[!hbt]
	\centering
	\caption{Ablation Studies on the LA Dataset.}
        \begin{threeparttable}
	\resizebox{0.45\textwidth}{!}{
	\begin{tabular}{cc|ccc|cc}
		\hline 
		\hline
		\multicolumn{2}{c|}{\% Scans Used}&\multicolumn{3}{c|}{Designs}&\multicolumn{2}{c}{Metrics}\\
		\hline
		Labelled&Unlabelled &IMD&MS&DiHC&Dice(\%)$\uparrow$ &ASD(voxel)$\downarrow$\\
		\hline
		\hline
		\multirow{6}{*}{10\%} &\multirow{6}{*}{90\%}
            & & & & 88.96 & 1.86 \\
            & & & \checkmark & & 88.69 & 1.84 \\
		& & & \checkmark & \checkmark & 89.09 & 1.69 \\
		& & \checkmark & & & 89.53 & 2.34 \\
		& & \checkmark & \checkmark & & 90.13 & 2.11 \\
		& & \checkmark & \checkmark & \checkmark & \textbf{90.42} & \textbf{1.54} \\
		\hline
		\multirow{6}{*}{20\%} &\multirow{6}{*}{80\%}
            & & & & 91.07 & 1.67 \\
            & & & \checkmark & & 91.11 & 1.5 \\            
		& & & \checkmark & \checkmark & 91.38 & 1.44 \\
		& & \checkmark & & & 91.1 & 1.82 \\
		& & \checkmark & \checkmark & & 91.72 & 1.72 \\
		& & \checkmark & \checkmark & \checkmark & \textbf{91.94} &\textbf{1.35} \\
		\hline
	\end{tabular}}
	\label{tabablation}
    \end{threeparttable}
\end{table}

We also conduct ablation studies to discern each proposed module's impact on the overall performance using LA dataset. We present the results to Tab. \ref{tabablation}. Our design components are abbreviated as: 1) Intra-Model Diversity (IMD), 2) Multi-scale Network (MS), and 3) Diagonal Hierarchical Consistency (DiHC). A baseline without any components is equal to \cite{wu2022mutual}. 

First, applying IMD does not consistently result in overall performance enhancement, particularly when comparing contour-related metric (ASD). This suggests that merely enhancing the diversity in the configurations of sub-models complicates the convergence process. However, the integration of MS and IMD results in performance improvement, as they fully leverage supervisory signals. Furthermore, integrating DiHC to the framework leads to significant gains across both metrics and training settings, verifying our initial hypothesis. 

\section{Conclusion}
\label{sec:conclusion}

In this paper, we present a concise but effective SSMIS framework. Its network is composed of three sub-models with the same multi-scale V-Net but distinct sub-layers. By employing deep supervision, they first learn the limited labelled data. Next, they collaborate by minimising differences on challenging regions by comparing soft pseudo labels from one's final predictions and others' intermediate and final outcomes via conventional mutual consistency and our diagonal hierarchical consistency on all data. Experimental results demonstrate the superiority and robustness of our DiHC-Net.

\bibliographystyle{IEEEtran}
\bibliography{refs}

\end{document}